\documentclass[10pt,twocolumn,letterpaper]{article}

\usepackage{cvpr}              

\usepackage{graphicx}
\usepackage{amsmath}
\usepackage{amssymb}
\usepackage{booktabs}
\usepackage{csquotes}
\usepackage{rotating}
\usepackage{pstricks}
\usepackage[normalem]{ulem}

\usepackage[pagebackref,breaklinks,colorlinks]{hyperref}

\usepackage{amsfonts,bm,mathtools,dsfont}
\def \vx{{\bm{x}}}
\def \vz{{\bm{z}}}
\def\gD{{\mathcal{D}}}

\usepackage[capitalize]{cleveref}
\crefname{section}{Sec.}{Secs.}
\Crefname{section}{Section}{Sections}
\Crefname{table}{Table}{Tables}
\crefname{table}{Tab.}{Tabs.}

\def\cvprPaperID{1682} 
\def\confName{CVPR}
\def\confYear{2022}

\begin{document}


\title{Dense Uncertainty Estimation via Ensemble-based Conditional Alternating Back-Propagation
}
\title{Dense Uncertainty Estimation via an Ensemble-based Conditional Latent Variable Model
}

\author{
Jing Zhang$^{1}$~
Yuchao Dai$^{2}$~
Mehrtash Harandi$^{3}$~ 
Yiran Zhong$^{4}$~ 
Nick Barnes$^{1}$~ 
Richard Hartley$^{1}$\\
$^1$ Australian National University\quad
$^2$ Northwestern Polytechnical University\\
$^3$ Monash University\quad
$^4$ SenseTime\\
}
\maketitle

\begin{abstract}
Uncertainty estimation has been extensively studied in recent literature, which can usually be classified as aleatoric uncertainty and epistemic uncertainty. In current aleatoric uncertainty estimation frameworks, it is often neglected that the aleatoric uncertainty is an inherent attribute of the data and can only be correctly estimated with an unbiased oracle model. Since the oracle model is inaccessible in most cases, we propose a new sampling and selection strategy at train time to approximate the oracle model for aleatoric uncertainty estimation. Further, we show a trivial solution in the dual-head based heteroscedastic aleatoric uncertainty estimation framework and introduce a new uncertainty consistency loss to avoid it. For epistemic uncertainty estimation, we argue that the internal variable in a conditional latent variable model is another source of epistemic uncertainty to model the predictive distribution and explore the limited knowledge about the hidden true model. We validate our observation on a dense prediction task, \ie, camouflaged object detection. Our results show that our solution achieves both accurate deterministic results and reliable uncertainty estimation.

\end{abstract}

\section{Introduction and Related Work}

\enquote{Uncertainty} represents our ignorance about the data or the model, which
can be roughly divided into aleatoric uncertainty (data uncertainty) and epistemic uncertainty (model uncertainty) \cite{kendall2017uncertainties}.
Effective uncertainty estimation can provide useful understanding about the limitations of the model, leading to explainable AI.
In this paper, we aim to extensively
explore uncertainty estimation techniques for dense prediction tasks in deep models for computer vision. We make important
observations about
undiscovered issues within the existing techniques, including the trivial solution and the hidden oracle assumption for aleatoric uncertainty. We also find the latent variable within a conditional latent variable model can act
as another origin of epistemic uncertainty. We conduct experiments on camouflaged object detection \cite{le2019anabranch,fan2020camouflaged}, a challenging dense prediction task. We show accurate deterministic results, and uncertainty prediction that is significantly more accurate than that from existing models.

\subsection{The source of uncertainty}
Let $\mathcal{D}$ be a  training set and consider a  predictive distribution
$p(y\vert \vx)$ defined as:
\begin{equation}
    \label{eqn:predictive_distribution}
    p(y|\vx) = \int p_\theta(y \vert \vx)q(\theta \vert \mathcal{D}) d\theta\;.
\end{equation}
Here, $p_\theta(y \vert \vx)$ is the likelihood of the model with parameters $\theta$,  and 
$q(\theta \vert \mathcal{D})$ is the approximate posterior for model.
For a generic multivariate regression task, $p_\theta(y|\vx)$ is usually modeled by a Gaussian distribution. 
In other words, $p_\theta(y \vert \vx)=\mathcal{N}(f_\theta(\vx),\Sigma)$, where $\Sigma$ is often a
diagonal matrix, indicating the inherent noise within the label. In a classification task, $p_\theta(y \vert \vx)$ is defined by a Softmax function, leading to $p_\theta(y \vert \vx)={\rm{Softmax}}(f_\theta(\vx)/T)$, with  $T>0$ being a temperature parameter, representing the uncertainty level~\cite{kendall2017uncertainties,on_calibration}.
Specifically, with the temperature, one can adjust the confidence of the model~\cite{on_calibration}.

\cref{eqn:predictive_distribution} shows that the randomness or uncertainty of $y$ comes from both inherent noise $\Sigma$ (or $T$ for classification tasks)
and $q(\theta|\gD)$, which is an approximation of the true distribution of model parameters $p(\theta)$. 
\cite{KIUREGHIAN2009105,kendall2017uncertainties}
define two types of uncertainty, namely aleatoric uncertainty \cite{aleatoric_epistemic_concept}, representing the intrinsic randomness of the task caused by $\Sigma$ (or $T$)
(\eg~labeling noise, sensor noise),
and epistemic uncertainty, caused by the gap between $q(\theta\vert \gD)$ and $p(\theta)$, indicating the limited knowledge about the task. The \enquote{intrinsic randomness} based aleatoric uncertainty usually cannot be reduced
with more data.
While, the \enquote{knowledge gap} based epistemic uncertainty can be explained away with more diverse training samples.
\subsection{Uncertainty modeling}


%
%


\noindent\textbf{Modelling aleatoric uncertainty:} Aleatoric uncertainty \cite{aleatoric_epistemic_concept,nicta_uncertainty,kendall2017uncertainties,Depeweg2018DecompositionOU,Collier2020ASP} captures the inherent noise, which can be decomposed into
input-independent uncertainty (homoscedastic uncertainty) and input-conditional uncertainty (heteroscedastic uncertainty). For the former, the \enquote{intrinsic randomness} of the data
is invariant to inputs. With a regression task, it is
assumed that the diagonal
covariance matrix $\Sigma$
is the same across all diagonal elements,
thus variance of the prediction is the same everywhere. In practice, one tunes the value of homoscedastic uncertainty instead of estimating it.
For the latter, the \enquote{intrinsic randomness} changes over inputs,
leading to varying
diagonal
covariance matrices
$\Sigma$, which in turn requires estimation.


Two main solutions exist for heteroscedastic aleatoric uncertainty estimation, including using an
auxiliary function based on a dual-head framework~\cite{nicta_uncertainty,kendall2017uncertainties}, and estimating
entropy or variance of multiple predictions with a stochastic prediction network, \eg~a Bayesian Neural Network (BNN) \cite{Depeweg2018DecompositionOU,quatification_uncertainty}. For the former, a dual-head framework is designed to produce both task related prediction $f_\theta(\vx)$ and its corresponding aleatoric uncertainty map $\sigma^2(\vx)$. For the latter, given a stochastic prediction network, the aleatoric uncertainty is defined as: $U_a = \mathbb{E}_{q(\theta \vert \gD)}[\mathbb{H}(y \vert \vx;\theta)]$~\cite{Depeweg2018DecompositionOU}, where $\mathbb{H}(\cdot)$ represents the entropy.
As $\mathbb{E}_{q(\theta \vert \gD)}$ is intractable, approximations are needed,
\eg,~Monte Carlo Dropout \cite{dropout_bayesian}, deep ensemble \cite{simple_scalable_uncertainty}.

\noindent\textbf{Modelling epistemic uncertainty:} Epistemic uncertainty \cite{Chitta2018DeepPE,aleatoric_epistemic_concept,Durasov_2021_CVPR,epistemic_ian,dropout_bayesian,random_relu,deterministic_epistemic,equipping_dnn_uncertainty,van2020uncertainty,kuppers2021bayesian,latent_derivative_bayesian} is assumed to be related to model parameters, capturing the limited knowledge about the model. To model epistemic uncertainty, a prior distribution is usually assigned over model parameters (\eg, $p(\theta)=\mathcal{N}(0,\mathbf{I})$). As the standard normal distribution may not be informative enough in some complicated cases, a
more sophisticated BNN is usually built to estimate the epistemic uncertainty
\cite{keep_nettwork_simple,bayesian_learning_for_nn,weight_uncertainty_nn,explaining_bnn,what_are_bayesian_looklike,aitchison2021a,global_inducing_point_variational}, where the
goal of the
BNN is to estimate the posterior predictive distribution.
Given
the dataset $\gD$, the
BNN aims to estimate the prediction $y^\ast$ of a new input $\vx^\ast$, which is described as the predictive posterior distribution:
\begin{equation}
    \label{bnn_posterior_predictive}
    p(y^\ast \vert \vx^\ast,\gD)=\int p_\theta(y^ \ast \vert \vx^\ast)q(\theta \vert \vx,y)d\theta,
\end{equation}
requiring a posterior over model parameters $q(\theta \vert \vx,y)$. According to
Bayes'
rule, $q(\theta \vert \vx,y)=p_\theta(y \vert \vx)\frac{p(\theta)}{p(y \vert \vx)}$. As the marginal likelihood $p(y \vert \vx)$ cannot be evaluated analytically, $q(\theta \vert \vx,y)$
has no closed-form solution, leading to
intractable posterior predictive distribution.
As a result,
approximations are required
using
two main solutions, namely
Variational Inference (VI) \cite{Jordan99anintroduction,Graphical_Models_VI,advance_variational_inference} and Markov Chain Monte Carlo (MCMC) \cite{mcmc_sampling,bridging_the_gap_mcmc}. The former approximates
the intractable posterior distribution $q(\theta \vert \gD)$ with an auxiliary tractable distribution $q_\phi(\theta)$ ($\phi$ is the variational parameter) by minimizing the Kullback–Leibler divergence between $q_\phi(\theta)$ and $q(\theta \vert \gD)$.
As a sampling based method, the latter draws a correlated sequence of $\theta_t$ from $q(\theta \vert \gD)$ ($t$ indicates the iteration of sampling)
to approximate the posterior predictive distribution as a Monte Carlo average. 
The BNN or its
approximation based uncertainty estimation methods \cite{weight_uncertainty_nn,dropout_bayesian,batchensemble,simple_scalable_uncertainty}
require multiple forward passes at test time, where
the epistemic uncertainty is
defined
as conditional mutual information between model prediction and model parameters: 
$U_e=\mathcal{I}[y,\theta \vert \vx, \gD] = \mathbb{H}[p(y \vert \vx, \gD)] - \mathbb{E}_{q(\theta \vert \gD)}[\mathbb{H}[y \vert \vx;\theta]]$. The compute burden at test time hinders their real-world applications.

Recently, some single-pass uncertainty estimation methods have been
introduced with
a deterministic network
\cite{van2020uncertainty,mukhoti2021deterministic,deterministic_epistemic,Liu2020SimpleAP,vanamersfoort2021feature}
or a stochastic network
\cite{Carvalho_2020_CVPR,sampling_free_variation_inference,Postels_2019_ICCV_Sampling_Free,lightweight_probabilistic_network}
.
The deterministic network
\cite{van2020uncertainty,mukhoti2021deterministic,deterministic_epistemic,Liu2020SimpleAP,vanamersfoort2021feature}
based methods
introduce a
distance-aware output representation to quantify the distance from
a test
example to
the training dataset manifold, where a test
example that is far
from the training data manifold is claimed to have higher epistemic uncertainty. However, these methods
may
suffer from feature collapse \cite{van2020uncertainty}, where the out-of-distribution samples are mapped to the same feature space as
in-distribution samples.
The stochastic network based single-pass epistemic uncertainty modeling methods \cite{Carvalho_2020_CVPR,lightweight_probabilistic_network} usually adopt
Gaussian Processes (GPs) \cite{Rasmussen2004} or variational inference estimation methods with strict constraints to obtain the evidence lower bound \cite{sampling_free_variation_inference} in closed form, which is
usually difficult to extend to high-dimensional data, \ie~images or videos.
\subsection{Uncertainty related applications}
Uncertainty is generally
used in three main contexts.
Firstly,
it is used as weight for loss or prediction
re-weighting \cite{kendall2018multi,Groenendijk_2021_WACV}, where the contribution of each part of the
loss function or each prediction is weighted based on the
certainty level of each task (multi-task learning) \cite{kendall2017uncertainties}, or each loss function (single task learning) \cite{Groenendijk_2021_WACV}, or each prediction \cite{Mirikharaji_2021_CVPR}.
Secondly, uncertainty is treated as
guidance for pseudo label quality estimation
for weakly/semi-supervised learning \cite{Yang_2021_CVPR,Nguyen_2021_ICCV}.
Thirdly, uncertainty serves as extra information in addition to
the task related prediction, leading to
error-awareness models for fully-supervised learning \cite{Truong_2021_CVPR,Upadhyay_2021_ICCV,Lu_2021_ICCV,S_2021_CVPR_domain,Zhuang_2021_CVPR,Li_2021_CVPR_ordinal_embedding,labelnotperfect}.
Although many
uncertainty related applications
\cite{Peng_2021_ICCV,Ekmekci_2021_ICCV,Collier_2021_CVPR,Chatterjee_2021_ICCV,LIU2021108140,Xu_2021_ICCV_mutltiview,generating_counterfactual,weakly_supervised_localization_uncertainty,multi_view_uncertainty,Well_Calibrated_Regression}
have been proposed, we notice they focus on directly using uncertainty without thoroughly
analysing the limitations of the uncertainty estimation techniques. We argue that a more comprehensive understanding about uncertainty estimation can further boost the uncertainty related applications.

\section{Observations and Contributions}
\label{sec:observations}

\subsection{Observations}
\noindent\textbf{Trivial solutions:}
Given dataset $\gD=\{\vx_i,y_i\}_{i=1}^N$ of size
$N$,
with the dual-head based aleatoric uncertainty from \cite{kendall2017uncertainties,nicta_uncertainty}, the loss function with a
noise corruption model for regression tasks
is defined as:
\begin{equation}
    \label{loss_aleatoric_uncertainty_sum}
    \mathcal{L}(\theta)=\frac{1}{N}\sum_{i=1}^N(\frac{1}{2\sigma^2(\vx_i)}\mathcal{L}(f_\theta(\vx_i),y_i)+\log\sigma(\vx_i)),
\end{equation}
and for classification tasks
it is written as:
\begin{equation}
\label{approximated_classification_log_likelihood_loss_sum}
    \mathcal{L}(\theta)=\frac{1}{N}\sum_{i=1}^N\frac{1}{T}\mathcal{L}(f_\theta(\vx_i),y_i) + \log T.
\end{equation}
In \cite{kendall2018multi}, the temperature is defined as $T=\sigma^2(\vx)$, and the derivation of \cref{approximated_classification_log_likelihood_loss_sum} is conditioned on $T \to 1$ (or $\sigma \to 1$). However, our
goal is to obtain accurate and confident predictions;
over-confident or under-confident
models should both be avoided. We intend to have $T=1$ most of the time, with  $T > 1$ when the model is less confident. As such, we define $T=\exp(\sigma^2(\vx))$ and condition \cref{approximated_classification_log_likelihood_loss_sum}  on $\sigma \to 0$ (or $T \to 1$). 
As there is no extra supervision
for estimating
aleatoric uncertainty in the dual-head framework, we observe one trivial solution,
which is $\sigma^2(\vx)=1$ for the regression task and 
and $\sigma^2(\vx)=0$ for the classification task.
The trivial solution indicates that the inherent noise ($\sigma^2$ in \cref{loss_aleatoric_uncertainty_sum} or $T$ in \cref{approximated_classification_log_likelihood_loss_sum})
is independent of the input, making it incapable of representing
heteroscedastic uncertainty.

\noindent\textbf{The hidden oracle assumption:} Recall that the noise corruption model \cite{nicta_uncertainty} for a regression task is defined as:
\begin{equation}
    \label{noise_corruption_model}
    y=f_\theta(\vx)+n(\vx), n(\vx)\sim\mathcal{N}(0,\Sigma).
\end{equation}
To exclude
model bias for aleatoric uncertainty modeling, the hidden assumption behind \cref{noise_corruption_model} is that $\theta$ should be unbiased, or it should be the \enquote{oracle model} or a perfect model \cite{deup_arxiv}.
However, this
is not reflected in the aleatoric uncertainty modeling strategy as in \cref{loss_aleatoric_uncertainty_sum}. Further, the mean entropy based aleatoric uncertainty \cite{Depeweg2018DecompositionOU} within a stochastic network also fails to reveal the \enquote{oracle assumption}. 


\noindent\textbf{Less comprehensive total uncertainty:} 
Uncertainty estimation models based on the BNN or its approximations define
predictive uncertainty \cite{Depeweg2018DecompositionOU} (the sum of aleatoric and epistemic uncertainty) as
entropy of the
mean model prediction.
Epistemic uncertainty is then defined as the mutual information between model prediction and model parameters. However, recall that
mutual information based uncertainty is defined only if the \enquote{entropy of the mean prediction} is a good representation of
total uncertainty. We argue that \enquote{total uncertainty} from the BNN or its variants is not sufficient
to explain the total uncertainty. For example, even if the training samples are generated from the true joint data distribution $p(\vx,y)$ (indicating low epistemic uncertainty), an
imbalanced sample distribution may still increase
total uncertainty. In this way, we argue that factors related to the
prediction distribution should also be included for total uncertainty estimation.


\noindent\textbf{Expensive
sampling process:} Although aleatoric uncertainty can be obtained with the dual-head framework \cite{nicta_uncertainty,kendall2017uncertainties}, both epistemic and predictive uncertainty usually rely on multiple forward passes to generate a sequence of predictions,
which is not
efficient when used at test time.

\subsection{Main contributions}
We intend to solve the above
limitations of existing uncertainty estimation methods with a single forward pass model for uncertainty-aware dense prediction tasks. Specifically, we investigate
ensemble-based \cite{simple_scalable_uncertainty} latent variable models \cite{ABP_aaai,gan_raw,VAE_Kingma}, where the
latent variable $\vz$ can be used to model the \enquote{predictive distribution}, leading to more comprehensive predictive uncertainty modeling.
From one view, the latent variable can capture the difficulty of labeling, meaning it
originates from
aleatoric uncertainty. In this paper, we define aleatoric uncertainty as uncertainty based on the \enquote{optimal prediction} within our latent variable model to explore the contribution of $\vz$ for aleatoric uncertainty with the \enquote{oracle constraint}. From another view, the latent variable aims to capture the important attributes of the task, which may sparsely exist in the provided training dataset. We define predictive uncertainty within our ensemble-based latent variable model to further explore the contribution of $\vz$ for epistemic uncertainty estimation.

Specifically, we extend the existing latent variable model (alternating back-propagation  (ABP) \cite{ABP_aaai}) to an
ensemble-based conditional alternating back-propagation (cABP), where the prior of the latent variable is assumed to be
conditioned
on the input, leading to both more efficient training and effective exploration of
latent space.
Note that
the widely studied
latent variable models either use an
extra inference model to approximate the true posterior distribution of the latent variable
(variational auto-encoder (VAE) \cite{VAE_Kingma,cvae}), or introduce a discriminator (generative adversarial network (GAN) \cite{NIPS2014_5423_gan}) to distinguish prediction and ground truth, where the latent variable is sampled from a fixed standard normal distribution.
\cite{ABP_aaai} samples $z$ directly from its true posterior distribution with gradient based Markov Chain Monte Carlo (MCMC) (Langevin dynamics \cite{mcmc_langevin} in particular), leading to more effective latent space modeling, avoiding the possible posterior collapse issue \cite{Lagging_Inference_Networks}.
Our cABP takes a step further by modeling the prior of the
latent variable as conditioned on the input instead of the less informative standard normal distribution, which is demonstrated to be
beneficial for both model training and latent space representation.


Within the proposed ensemble-based cABP framework, we solve the \enquote{trivial solution} problem with an uncertainty consistency loss. We solve the \enquote{hidden oracle assumption} by defining the aleatoric uncertainty as data uncertainty based on the optimal prediction, which can be defined as an approximation of prediction from the oracle model. The latent variable is introduced
to a conditional latent variable model solve the \enquote{less comprehensive total uncertainty} issue. Further, to produce uncertainty with a single forward pass, we
design an uncertainty distillation network to achieve sampling-free uncertainty at test time.


\section{Our Method}
\label{sec:our_method}
We aim to build an effective and general uncertainty estimation network
to produce both accurate prediction and reliable uncertainty as shown in Fig.~\ref{fig:model_overview}.

\subsection{Ensemble-based latent variable model}
Deep ensemble \cite{simple_scalable_uncertainty} has been shown to be
an effective epistemic uncertainty estimation technique, which initializes the network or part of the network differently to produce multiple predictions.
To avoid the expensive memory requirement, we build a last-layer ensemble-based framework with $M=5$ ensembles, which is the cheapest decoder ensemble.
As discussed, the latent variable within the latent variable model \cite{VAE_Kingma,cvae,gan_raw} can capture both the labeling noise and the
predictive distribution.
In this paper, we combine the deep decoder ensemble and the latent variable model to capture comprehensive epistemic uncertainty.
We introduce conditional ABP \cite{ABP_aaai},
where prior of the latent variable is conditioned on the input image.





\noindent\textbf{Alternating back-propagation:}
ABP \cite{ABP_aaai} was introduced for learning the generator network model. It updates the latent variable and network parameters in an EM-manner. Firstly, given the network prediction with the current parameter set, it infers the latent variable by
Langevin dynamics based MCMC \cite{mcmc_langevin}, which \cite{ABP_aaai}
calls
\enquote{Inferential back-propagation}.
Secondly, given the updated latent variable $z$, the network parameter set is updated with gradient descent, called
\enquote{Learning back-propagation}
\cite{ABP_aaai}.
Following the previous variable definitions,
given the training example $(\vx,y)$, we intend to infer $\vz$ and learn the network parameter $\theta$ to minimize the reconstruction error as well as a regularization term that corresponds to the prior on $\vz$.

A typical conditional generative model is defined as:
\begin{eqnarray}
    && \vz \sim q(\vz)=\mathcal{N}(0,\mathbf{I}), \label{eq:abp_1}\\
    && y = f_\theta(\vx,\vz) + \epsilon, \epsilon \sim \mathcal{N}(0,\Sigma).
    \label{eq:abp_3} 
\end{eqnarray}
$q(\vz)$ is the prior distribution of $\vz$, and $\Sigma = \text{diag}(\sigma^2)$ is the inherent label noise. The conditional distribution of $y$ given $\vx$ is $p_\theta(y|\vx) = \int q(\vz) p_\theta(y|\vx,\vz) d\vz$.
The observed-data log-likelihood is then defined as $L(\theta)=\sum_{i=1}^N \log p_\theta(y_i|\vx_i)$, where the
gradient of $p_\theta(y\vert \vx)$ can be obtained via:
\begin{equation}
\label{update_omega}
    \frac{\partial}{\partial \theta}\log p_\theta(y \vert \vx)=\text{E}_{p_\theta(\vz \vert \vx,y)} \left[\frac{\partial}{\partial \theta}\log p_\theta(y,\vz \vert \vx)\right].
\end{equation}

\begin{figure}[ht!]
   \begin{center}
   \begin{tabular}{c@{ }}
\includegraphics[width=0.95\linewidth]{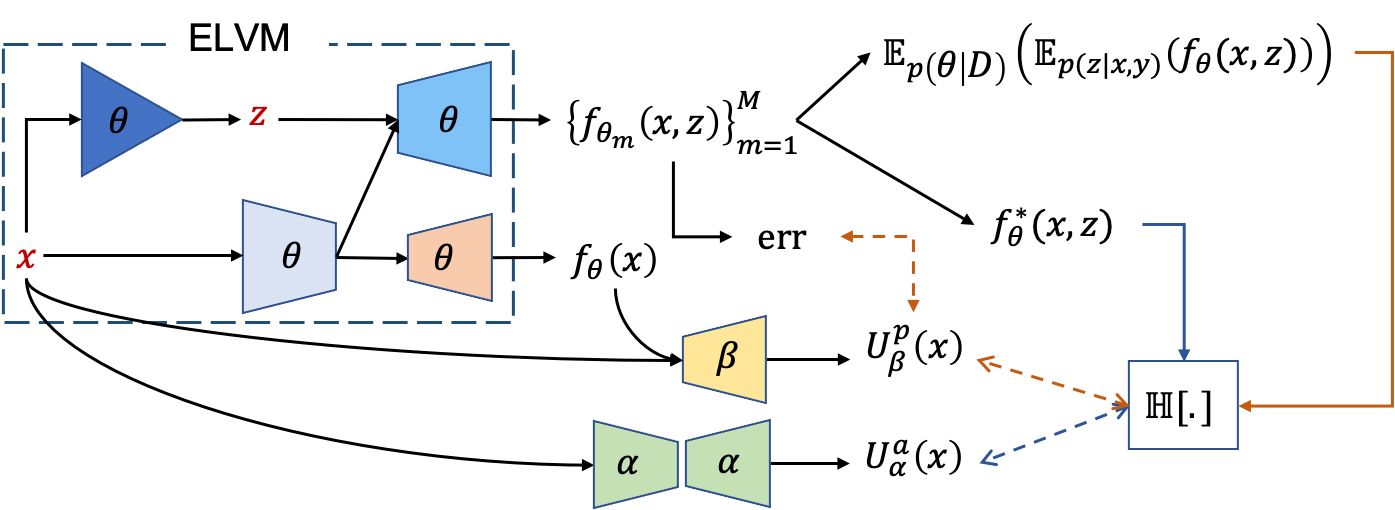}\\
\end{tabular}
   \end{center}
   \caption{Pipeline of the proposed single-pass ensemble-based latent variable model (ELVM)
   for uncertainty estimation,
   where $\theta$, $\alpha$ and $\beta$ represent the parameters of our ELVM, the aleatoric and predictive uncertainty estimation module respectively.}
\label{fig:model_overview}
\end{figure}

We approximate the expectation term $\text{E}_{p_\theta(\vz \vert \vx,y)}$
by drawing samples from $p_\theta(\vz \vert \vx,y)$.
Following ABP \cite{ABP_aaai}, we use a gradient-based Monte Carlo method, namely Langevin Dynamics \cite{mcmc_langevin},
to sample $\vz$:
\begin{equation}
\begin{aligned}
    \vz_{t+1}= \vz_{t}+ \frac{s^2}{2}\left[ \frac{\partial}{\partial \vz}\log p_\theta(y,\vz_{t} \vert \vx)\right]+s \mathcal{N}(0,\mathbf{I}),
    \label{langevin_dynamics}
\end{aligned}
\end{equation}
where the gradient term is defined as: 
\begin{equation}
\frac{\partial}{\partial \vz}\log p_\theta(y,\vz \vert \vx) = \frac{1}{\sigma^2}(y-f_\theta(\vx,\vz))\frac{\partial}{\partial \vz}f_\theta(\vx,\vz) - \vz\;,
\end{equation}
where $t$ is the
time step for Langevin sampling, and $s$ is the step size. \cref{langevin_dynamics} shows an effective latent variable inference model by
directly sampling it from the true posterior distribution $p_\theta(\vz \vert \vx,y)$.

Different
from VAEs
\cite{VAE_Kingma,cvae}, where an
extra inference model $q_\phi(\vz \vert \vx,y)$ with
parameters $\phi$ is designed to approximate the true posterior distribution of the latent variable, and GANs
\cite{gan_raw}, where the discriminator is used to distinguish prediction and ground truth, ABP
\cite{ABP_aaai} samples the latent variable directly from its true posterior distribution via Langevin Dynamic based MCMC \cite{mcmc_langevin} with no extra inference model involved.

\noindent\textbf{Conditional Alternating Back-propagation:} Although the posterior of $\vz$ is sampled directly from its true posterior distribution via \cref{langevin_dynamics},
directly applying
ABP \cite{ABP_aaai} to the
dense prediction task may still be problematic, as the prior of $\vz$ is assumed to follow the input-independent standard normal distribution with $q(\vz)=\mathcal{N}(0,\mathbf{I})$.
To obtain an
image-conditioned prior, we introduce conditional ABP, where the
prior of $\vz$ is modeled with an extra encoder as shown in \cref{fig:model_overview}. Let's define the prior of $\vz$ as $\vz_0\sim p_\theta(\vz \vert \vx)$, and model it with a convolutional neural network. Taking it back to \cref{langevin_dynamics}, we obtain our conditional ABP.


\subsection{Fixing the limitations of existing uncertainty estimation techniques}

\noindent\textbf{Avoiding
the trivial solutions with the
hidden oracle assumption:}
The basic assumption for
dual-head based aleatoric uncertainty is that
the
model output $f_\theta(\vx)$ should be an unbiased oracle.
However, this hidden assumption
is not indicated in the related loss function in Eq.~\ref{loss_aleatoric_uncertainty_sum} or Eq.~\ref{approximated_classification_log_likelihood_loss_sum}. In this paper, we define the aleatoric uncertainty explicitly based on the assumption
within a multi-head framework as shown in Fig.~\ref{fig:model_overview}. To achieve this, with multiple predictions $\{f_{\theta_m}(\vx,\vz)\}_{m=1}^M$ from the ensemble-based latent variable model, we first compute the
pixel-wise loss of each prediction. Then, we generate a new prediction $f^\ast_\theta(\vx,\vz)$ as the optimal prediction, where each pixel of $f^*_\theta(\vx,\vz)$ has the least loss. This \enquote{optimal prediction} is assembled from the models, and we treat it as an approximation to the oracle.
Finally, we
define its uncertainty as the aleatoric uncertainty.
For a given sample $\vx$, its aleatoric uncertainty is irreducible. \cite{deup_arxiv} defines aleotoric uncertainty
as being estimated
by a Bayes-optimal predictor. Similarly, in dense prediction tasks, we define the aleatoric uncertainty as the uncertainty of the optimal prediction
to approximately reveal the \enquote{hidden oracle} assumption. To avoid the trivial solution, we introduce an uncertainty consistency loss with a multi-head aleatoric uncertainty framework,
where the consistency loss is defined between the aleatoric uncertainty from the multi-head framework ($U^a_\alpha(\vx)$) and the uncertainty of the optimal prediction ($\mathbb{H}[f^\ast_\theta(\vx,\vz)]$).

\noindent\textbf{Latent variable as origin of predictive uncertainty:}
The predictive distribution of a Bayesian latent variable model is defined as: $p(y|\vx)=\int p_\theta(y \vert \vx,\vz)q(\theta \vert \mathcal{D}) q(\vz|\gD) d\vz d\theta\;$. \cite{Depeweg2018DecompositionOU} defines $p(\vz|\gD)=p(\vz)$, indicating an input-independent latent variable, and they then define $\vz$ as another source of aleatoric uncertainty, which is consistent with the input-independent latent variable assumption.
However, in the conditional case, the latent variable models $p_\theta(\vz|\vx,y)$. This includes the mapping from input $\vx$ to output $y$ by $\theta$. That is, $\vz$ models predictive uncertainty, including both epistemic and aleotoric uncertainty. In this paper, instead of defining $\vz$ as the source of aleatoric uncertainty, we claim it is also the origin of epistemic uncertainty in the conditional case, making it the origin of predictive uncertainty in general.

\noindent\textbf{Single-pass uncertainty with uncertainty distillation:} To achieve efficient uncertainty estimation at
test time, we introduce a single-pass uncertainty estimation framework with an
uncertainty consistency loss as shown in Fig.~\ref{fig:model_overview}. Specifically, we want to obtain aleatoric uncertainty $U^a_\alpha$ and predictive uncertainty $U^p_\beta$ directly, and the epistemic uncertainty $U^e$ is then defined as $U^e=U^p_\beta-U^a_\alpha$.
For the aleatoric uncertainty, to avoid the trivial solution and also to maintain the \enquote{oracle assumption}, we generate the
optimal prediction, and define the uncertainty consistency loss between $U^a_\alpha$ and the uncertainty of the optimal prediction.
For the predictive uncertainty, its basic definition is $U^p= \mathbb{H}[\mathbb{E}_{p(\theta \vert \gD)}(\mathbb{E}_{p(\vz \vert \vx,y)}(f_\theta(\vx,\vz)))]$, meaning that the predictive uncertainty is the uncertainty of the
mean model prediction within an ensemble-based latent variable model.
In addition, we also claim that the predictive uncertainty should represents the overall awareness of model towards its prediction. In this way, the predictive uncertainty should highlight the error predictions. We then define the second part of the predictive uncertainty as the mean error: $\text{err}=(\mathbb{E}_{p(\theta \vert \gD)}(\mathbb{E}_{p(\vz \vert \vx,y)}(f_\theta(\vx,\vz)-y))^2$.

\noindent\textbf{Deterministic prediction generation:} For the existing multiple forward passes based uncertainty estimation models, an expensive
sampling process is needed to generate the multiple predictions, and then the mean prediction is used as the deterministic prediction at test time. For our single-pass based uncertainty estimation models, we introduce an
extra deterministic prediction branch ($f_\theta(\vx)$ in Fig.~\ref{fig:model_overview}) without any randomness involved. The main focus of this branch is to generate efficient deterministic prediction at test time.

\subsection{Training the model}
Given \cref{loss_aleatoric_uncertainty_sum,approximated_classification_log_likelihood_loss_sum}, in practice, we do not regress $\sigma(\vx)$ directly. Instead, following \cite{kendall2017uncertainties}, we define $s(\vx)=\log(\sigma^2(\vx))$ to achieve stable training. As shown in \cref{fig:model_overview}, we have two main loss functions within our framework, namely the \enquote{Task related loss} and the uncertainty consistency loss.
The \enquote{Task related loss} for the
regression task
within both the deterministic model (the $f_\theta(\vx)$ branch) and the stochastic model ($f_\theta(\vx,\vz)$) in \cref{fig:model_overview}) is defined as:
\begin{equation}
    \label{deterministic_pred_loss}
    \mathcal{L}_d(\theta)=\frac{1}{N}\sum_{i=1}^N(\frac{1}{2\sigma^2(\vx_i)}\mathcal{L}(f_\theta(\vx_i),y_i)+\log\sigma(\vx_i)),
\end{equation}
\begin{equation}
    \label{stocastic_pred_loss}
    \mathcal{L}_s(\theta)=\frac{1}{N}\sum_{i=1}^N(\frac{1}{2\sigma^2(\vx_i)}\mathcal{L}(f_\theta(\vx_i,\vz_i),y_i)+\log\sigma(\vx_i)),
\end{equation}
where $\sigma^2(\vx)$ is the estimated predictive uncertainty, and $\mathcal{L}$ is the chosen loss function.
The \enquote{Task related loss} for the
dense classification task in \cref{approximated_classification_log_likelihood_loss_sum} can be extended similarly.



For the uncertainty estimation modules $U^a_\alpha(\vx)$ and $U^p_\beta(x,f_\theta(\vx,\vz))$, we define uncertainty consistency loss functions to avoid the trivial solution and also achieve single-pass uncertainty estimation at test time. 

The aleatoric uncertainty consistency loss is defined as:
\begin{equation}
    \label{aleatoric_uncertainty_consistency}
    \mathcal{L}_{au} = \mathcal{L}_{bice}(U^a_\alpha(\vx),\mathbb{H}[f^*_\theta(\vx,\vz)]),
\end{equation}
where $\mathcal{L}_{bice}$ is the bi-directional cross-entropy loss.
For predictive uncertainty, with the mean prediction 
$p_\theta(y\vert \vx,\vz)=\mathbb{E}_{p(\theta \vert \gD)}(\mathbb{E}_{p(\vz \vert \vx,y)}(f_\theta(\vx,\vz))$ and mean error $\text{err}=(\mathbb{E}_{p(\theta \vert \gD)}(\mathbb{E}_{p(\vz \vert \vx,y)}(f_\theta(\vx,\vz)-y))^2$,
the uncertainty consistency loss is then defined as:
\begin{equation}
    \label{predictive_uncertainty_consistency}
    \begin{aligned}
        \mathcal{L}_{pu} =& \mathcal{L}_{bice}(U^p_\beta(\vx,f_\theta(\vx,\vz)),\mathbb{H}[p_\theta(y \vert \vx,\vz)])\\
        &+\mathcal{L}_{ce}(U^p_\beta(\vx,f_\theta(\vx,\vz)),\text{err}),
    \end{aligned}
\end{equation}
where
$\mathcal{L}_{ce}$ is the binary cross-entropy loss. Note that, we perform min-max normalization for the target of the bi-directional cross-entropy loss and the cross-entropy loss. As discussed, the output of each uncertainty module is $s=\log (\sigma^2(\vx))$, and $U^a_\alpha$ and $U^p_\beta$ are then defined as $\exp{(s)}$. Similarly, we perform min-max normalization before we feed them to the corresponding loss functions.

\noindent\textbf{Updating $\theta$:} Within our single-pass ensemble-based latent variable model, we have $M=5$ ensembles, leading to
$\theta=\{\theta_m\}_{m=1}^M$.
We use the two task related loss functions to update $\theta$, which is: $\mathcal{L}_{task}=\mathcal{L}_d(\theta)+\mathcal{L}_s(\theta)$. Note that, the prior of the latent variable $p_\theta(\vz \vert \vx)$ is updated simultaneously with $\mathcal{L}_{task}$. 

\noindent\textbf{Updating $\alpha$ and $\beta$:} The aleatoric uncertainty estimation module takes image $\vx$ as input, and outputs $U^a_\alpha(\vx)$. $\alpha$ is then updated with loss function in \cref{aleatoric_uncertainty_consistency}. The predictive uncertainty estimation module takes the image $\vx$ and a stochastic prediction $f_\theta(\vx,\vz)$ as input, and
is updated with the
loss function in \cref{predictive_uncertainty_consistency}.



\subsection{Network details}
\noindent\textbf{The ensemble-based latent variable model:} We take the
ResNet50 backbone \cite{he2016deep} as our encoder for the ensemble-based latent variable model (ELVM), which maps the input image $\vx$ to backbone features $\{s_l\}_{l=1}^4$ of channel size 256, 512, 1024 and 2048 respectively. To reduce the
channel size with larger receptive fields, we feed each $s_l$ to
multi-scale dilated convolutional layers \cite{denseaspp} to obtain new backbone features $\{s'_l\}_{l=1}^4$ of consistent channel size $C=32$. Then, we have a stochastic prediction module with an
ensemble structure to produce $\{f_{\theta_m}(\vx,\vz)\}_{m=1}^M$, and a deterministic prediction module to generate $f_\theta(\vx)$. Both
take $\{s'_l\}_{l=1}^4$ as input, and the decoder of each $f_{\theta_m}(\vx,\vz)$ is the same as $f_\theta(\vx)$. This
is borrowed from \cite{midas_tpami}, except that, for the stochastic prediction module, we concatenate $\vz$ with $s'_4$, and feed it to another convolutional layer of kernel size $3\times3$ to obtain the new $s'_4$. Note that as a last-layer ensemble based framework, we attach $M=5$ copies of the last layer $\{\text{cls}_m\}_{m=1}^M$ in the stochastic prediction module.


Within ELVM, we introduce an
extra encoder $p_\theta(\vz \vert \vx)$ (the prior model) to model the prior distribution of the latent variable $\vz$, leading to $\vz_0\sim p(\vz \vert \vx)$. Specifically, the prior model consists of five convolutional layers and two fully connected layers. The convolutional layers have the same
kernel size of
$4\times4$, stride size of
2 and output channel size of $C=32$. The two fully connected layers are adopted to obtain mean $\mu$ and variance $\sigma^2$ respectively, and the latent variable $\vz$ is then obtained with the reparameterization trick \cite{VAE_Kingma}: $\vz=\mu+\sigma\odot\epsilon$, where $\epsilon\sim\mathcal{N}(0,\mathbf{I})$. We set dimension of the latent variable $z$ as $K=32$ in our framework.


\noindent\textbf{The uncertainty estimation modules}: The aleatoric uncertainty estimation modules $U^a_\alpha(\vx)$ takes only image $\vx$ as input.
Empirically,
we design it also with a
ResNet50 backbone \cite{he2016deep} as our encoder, and a
decoder from \cite{midas_tpami}.
As the predictive uncertainty estimation module $U^p_\beta(\vx,f_\theta(\vx,\vz))$ also takes the prediction as input, we then design a structure without a backbone network. Specifically, $U^p_\beta(\vx,f_\theta(\vx,\vz))$ 
consists five convolutional layers of the same kernel size $3\times3$. The stride sizes of the five convolutional layers are 2, 1, 2, 1, 2 respectively. The first four convolutional layers have channel size of 64, and the last convolutional layer produces a one-channel feature map. We then define its power exponent as the corresponding uncertainty \cite{kendall2017uncertainties}. Note that,
we use LeakyReLU and batch normalization after the first four convolutional layers.

\section{Experimental Results}
\label{sec:experiments}
To verify our general pipeline for single-pass uncertainty estimation, we perform experiments on a dense prediction tasks, namely camouflaged object detection (COD) \cite{le2019anabranch,fan2020camouflaged}.

\subsection{Uncertainty within the task
}
Camouflaged objects are
objects that hide in the environment, which usually share similar appearance to
their surroundings.
Camouflaged object detection \cite{fan2020camouflaged,le2019anabranch}
suffers greatly from both inherent noise related uncertainty and uncertainty related to the
knowledge gap arising from
the definition of
camouflage.
From the labeling point of view, due to similar patterns
of camouflaged objects and their surroundings,
it is difficult to precisely locate and depict camouflaged objects, leading to inherent noise within the labeling process conducted by annotators.
On the other hand, in realistic scenes, there should not exist any location related prior for camouflaged objects. 
However, as
existing COD training datasets (\eg, \cite{le2019anabranch,fan2020camouflaged}) are collected from the Internet where the photographers focus on the camouflaged instance, there is a
serious \enquote{center bias} issue where most of the camouflaged instances are
located near the center of the image. Further, as a class-agnostic binary segmentation task, the
training dataset contains
limited categories of camouflaged objects, which may also lead to category-bias, where the model learns
to segment some specific category of objects, leading to category-aware learning.
%



\subsection{Experimental setup}
\noindent\textbf{Dataset:} The benchmark training dataset is the combination of 3,040 images from COD10K training dataset \cite{fan2020camouflaged} and 1,000 images from CAMO training dataset \cite{le2019anabranch}. The testing datasets include
CAMO testing dataset \cite{le2019anabranch}, CHAMELEMON dataset \cite{Chameleon2018},
COD10K testing dataset \cite{fan2020camouflaged},
and
NC4K dataset \cite{yunqiu2021ranking}.

\noindent\textbf{Evaluation Metrics:}
We produce both
deterministic predictions and
uncertainty maps representing the
model's limitations.

For deterministic prediction evaluation, we adopt the Mean Absolute Error and Mean F-measure denoted as $\mathcal{M}$ and $F_\beta$ respectively.

\textbf{MAE $\mathcal{M}$} is defined as per-pixel wise difference between model prediction $s$ and a per-pixel wise binary ground-truth $y$:
\begin{equation}
    \begin{aligned}
    \text{MAE} = \frac{1}{H\times W}|s-y|,
    \end{aligned}
\end{equation}
where $H$ and $W$ are height and width of $s$. 
MAE provides a direct
estimate of conformity between estimated and ground-truth maps.
However, for the MAE metric, small objects naturally assign a smaller error and
larger objects have larger errors.

\textbf{F-measure $F_{\beta}$} is a region based similarity metric, and we provide the mean F-measure using varying fixed (0-255) thresholds.

We use two evaluation metrics to measure uncertainty: 1) dense expected calibration error \cite{on_calibration,zhang2020uncertaintyaware} ($\text{ECE}_d$),
and
2)
patch accuracy vs patch uncertainty ($\text{PAvPU}$) \cite{evaluation_uncertainty}.

\textbf{Dense expected calibration error} $\text{ECE}_d$ \cite{on_calibration,zhang2020uncertaintyaware} aims evaluate model calibration degree, or in particular the gap between model accuracy and model confidence. Let's define model accuracy $\mathrm{acc}$ and model confidence $\mathrm{conf}$.
For model $f_\theta$ with parameters $\theta$ and a given testing set $G=\{\vx_i,y_i\}^S$ of size $S$, $\mathrm{acc}$ is used to measure the accuracy of the model on $G$. $\mathrm{conf}$ measures how much the model believes in its predictions, and $\text{ECE}_d$ measures the calibration error of model $f_\theta$ on $G$.

For an image $\vx_i$, we define its prediction according to $f_\theta$ as $s_i=f_\theta(\vx_i)$. Following~\cite{on_calibration}, we group predictions $s_i$ into $M$ interval bins\footnote{We set $M=12$ in our experiments, with the first and last bin containing predictions of $s_i^{(u,v)} = 0$ and $s_i^{(u,v)} = 1$, respectively.}. The accuracy of each bin is measured as:
\begin{equation}
\label{ori_accuracy}
\mathrm{acc}(B_m)^i=\frac{1}{|B_m|}\sum_{(u,v)\in B_m}\mathbf{1}(g(s_i^{(u,v)})=y_i^{(u,v)}),
\end{equation}
where $B_m$ {are} the samples that fall in the $m$-th interval bin, $|B_m|$ is the cardinality of $B_m$, $(u,v)$ represents coordinate of pixels in $B_m$, $g(.)$ is thresholding operation to transfer gray prediction to binary image. 

As our prediction $s_i$ is a gray scale image, we follow the idea of F-measure, and obtain a binary prediction by thresholding the map $s_i$ in the range of $[0,1]$ with 256 intervals.
Each $g(s_i^{(u,v)})$ lead to one accuracy as Eq. \eqref{ori_accuracy}. With 256 thresholds, we obtain a 256-d vector for accuracy $\mathrm{acc}(B_m)^i$ of each bin. $\mathrm{macc}(B_m)^i$ is defined as mean of $\mathrm{acc}(B_m)^i$. Then, the accuracy $\mathrm{acc}^i$ of image $\vx_i$ is defined as: $\mathrm{acc}^i = \{\mathrm{macc}(B_1)^i,\cdot,\mathrm{macc}(B_M)^i\}$, which is a $M$ dimensional vector, with each position representing accuracy for a specific bin $B_m$.

The prediction confidence for each pixel represents how much the model trusts its predictions. For an image $\vx_i$, the average confidence of each bin $B_m$ is then defined as:
\begin{equation}
    \mathrm{conf}(B_m)^i=\frac{1}{|B_m|}\sum_{(u,v)\in B_m} \hat{p}_i^{(u,v)},
\end{equation}
where $\hat{p}_i^{(u,v)}$ is the model confidence at position $(u,v)$, which is defined as:
\begin{equation}
    \hat{p}_i^{(u,v)} = \max\{s_i^{(u,v)},(1-s_i^{(u,v)})\},
\end{equation}

Dense expected calibration error $\text{ECE}_d$ is the weighted average of the difference between bins' accuracy and confidence. For image $\vx_i$, we define its dense expected calibration error as:
\begin{equation}
    \text{ECE}_d^i = \sum_{m=1}^M\frac{|B_m|}{\sum_m |B_m|}|\mathrm{macc}(B_m)^i-\mathrm{conf}(B_m)^i|,
\end{equation}
where $\sum_m |B_m|$ is the number of pixels in $\pmb{x}_i$. A perfectly calibrated model should have $\mathrm{conf}(B_m)^i = \mathrm{macc}(B_m)^i$, thus leads to $\text{ECE}_d^i=0$.
For a given testing dataset $G$ and trained model $f_\theta$, we define the dense expected calibration error of the model on $G$ as: $\text{ECE}_d^{G} = \mathrm{mean}\{\text{ECE}_d^1,\cdot,\text{ECE}_d^S\}$. $\text{ECE}_d^{G}=0$ represents the model is perfectly calibrated, and $\text{ECE}_d^{G}=1$ indicates poorly calibrated model. 

\textbf{Patch accuracy vs patch uncertainty} $\text{PAvPU}$ \cite{evaluation_uncertainty} is based on one assumption that for certain pixels (or regions), the model should have accurate predictions and for inaccurate predictions, the corresponding certain level should be low.

For each patch $s_k$, \cite{evaluation_uncertainty} computes the patch accuracy $a_k$ (pixel accuracy metric defined in \cite{Long_2015_CVPR}). Then, \cite{evaluation_uncertainty} decides the patch as accurate based on certain threshold $h_a$, \eg prediction of $s_k$ is accurate if $a_k > h_a$.
Similarly, the uncertainty of patch $s_k$ is achieved if it's uncertainty $u_k>h_u$, where $h_u$ is a confidence related threshold. They define $n_{ac}$ as number of patches which are accurate and certainty, $n_{au}$ as number of patches which are accurate and uncertainty, $n_{ic}$ as number of patches which are inaccurate and certainty and $n_{iu}$ as number of patches which are inaccurate and uncertainty. 
Then, the patch accuracy vs patch uncertainty (PAvPU) is defined as:
\begin{equation}
    \text{PAvPU} = \frac{(n_{ac}+n_{iu})}{(n_{ac}+n_{au}+n_{ic}+n_{iu})},
\end{equation}
and a model with higher above scores has better performer.

Instead of define uniform patch, we over-segment image to super-pixels to achieve uniform accuracy/uncertainty with in a same super-pixel considering the semantic consistency attribute. Further, the three above scores are based on handcrafted hyper-parameters, namely accuracy related threshold $h_a$ and uncertainty related threshold $h_u$. Inspired by the expected calibration error \cite{on_calibration}, we use bin instead of hard-thresholding. Specifically, we use threshold in the rang of $[0,1]$ with 10 bins, and the final $n_{ac}$, $n_{au}$, $n_{ic}$, $n_{iu}$ are all 10-dimensional vectors, as well as $\text{PAvPU}$. We then define mean $\text{PAvPU}$ for uncertianty performance evaluation. 

\begin{table*}[t!]
  \centering
  \scriptsize
  \renewcommand{\arraystretch}{1.2}
  \renewcommand{\tabcolsep}{1.3mm}
  \caption{Performance comparison with alternative uncertainty estimation methods .
  $\uparrow$ indicates the higher the better, and vice versa for $\downarrow$.}
  \begin{tabular}{l|cccc|cccc|cccc|cccc}
  \toprule
  Method &\multicolumn{4}{c|}{CAMO~\cite{le2019anabranch}}&\multicolumn{4}{c|}{CHAMELEMON~\cite{Chameleon2018}}&\multicolumn{4}{c|}{COD10K~\cite{fan2020camouflaged}}&\multicolumn{4}{c}{NC4K~\cite{yunqiu2021ranking}} \\
    &$F_{\beta}\uparrow$&$\mathcal{M}\downarrow$&$\text{ECE}_d\downarrow$&$\text{PAvPU}\uparrow$
    &$F_{\beta}\uparrow$&$\mathcal{M}\downarrow$&$\text{ECE}_d\downarrow$&$\text{PAvPU}\uparrow$
    &$F_{\beta}\uparrow$&$\mathcal{M}\downarrow$&$\text{ECE}_d\downarrow$&$\text{PAvPU}\uparrow$
    &$F_{\beta}\uparrow$&$\mathcal{M}\downarrow$&$\text{ECE}_d\downarrow$&$\text{PAvPU}\uparrow$\\
  \hline
  Base & .736 & .083 & .066 & .838 & .832 & .033 & .020 & .875 & .710 & .038 & .029 & .876 & .785 & .053  & .041 & .851  \\
  MD & .750 & .084 & .065 & .840 & .825 & .033 & .023 & .877  & .714 & .037 & .028 &  .880& .794 & .050 & .039 &.857 \\
  DE  & .749 & .085 & .063 & .842 & .827 & .035 & .021 & .878 & .711 & .038 & .027 &.888  & .794 & .051 & .038 &.857 \\
  GAN & .744 & .084 & .065 & .845 & .835 & .032 & .022 & .878  & .711 & .037 & .026 &.887  & .790 & .051 & .037 &.859 \\ 
  VAE & .738 & .086 & .065 & .837 & .827 & .033 & .023 &.876  & .712 & .037 & .028 &.878  & .792 & .050 & .039 &.854 \\
  \hline
  Ours & .739 & .082 & .044 & .902 & .830 & .034 & .011 & .937 & .713 & .037 & .013 & .929 & .780 & .054 & .024 & .913\\
  \bottomrule
  \end{tabular}
  \label{tab:alternative_uncertainty_methods}
\end{table*} 

\section{Alternative Generative Models}

As a latent variable model, we also compare with alternative latent variable models, namely generative adversarial net (GAN) \cite{gan_raw} and variational auto-encoder (VAE) \cite{VAE_Kingma,cvae}. Specifically, we define the
dimension of the latent variable $z$ same as ours within the
\enquote{GAN} and \enquote{VAE}.
For the GAN based model, an
additional discriminator is designed to distinguish the prediction and ground truth. For the VAE based framework, we use an
extra inference model $p(\vz \vert \vx,y)$ to approximate the true posterior distribution, and define $p(\vz \vert \vx)$ as the
prior of the latent variable following \cite{jing2020uc}. Similarly, we can obtain the three types of uncertainty with 10 forward passes at test time.

\noindent\textbf{Implementation details:} We trained the model
using PyTorch.
The ResNet50 \cite{ResHe2015} backbone for both the ensemble based latent variable model and the aleatoric uncertainty estimation module are initialized with parameters trained for image classification, and the newly added layers are initialized by the default initialization policy in Pytorch.
We adopt the structure-aware loss function \cite{wei2020f3net} as the \enquote{chosen loss function} in \cref{deterministic_pred_loss,stocastic_pred_loss}.
We use the Adam method with momentum 0.9 and decrease the learning rate by
10\% after 80\% of the maximum epochs. The base learning rate is initialized as 2.5e-5. The training takes around 15 hours with training batch size 22, and maximum epochs of 50 on a PC with an NVIDIA GeForce RTX 3090 GPU.

\subsection{Dense uncertainty estimation}
We compare the proposed single-pass ELVM with existing uncertainty estimation techniques, namely MC-dropout \cite{dropout_bayesian} \enquote{MD} and deep ensemble \cite{simple_scalable_uncertainty} \enquote{DE}. For the MC-dropout \cite{dropout_bayesian} technique, we add dropout with dropout rate=0.3 after each backbone features $\{s_l\}_{l=1}^4$. We keep the dropout layer at test time, and perform 10 forward passes to obtain the predictive uncertainty as the
entropy of the mean prediction, aleatoric uncertainty as the
mean of the entropy of the ten stochastic predictions. Epistemic uncertainty is then defined as the residual of predictive uncertainty and aleatoric uncertainty. For the
deep ensemble, we add five decoders (same as our deterministic decoder) to the backbone network, and the three types of uncertainty are
then obtained similarly to
MC-dropout.

\begin{figure*}[ht!]
  \begin{center}
  \begin{tabular}{c@{ }}
\includegraphics[width=1.0\linewidth]{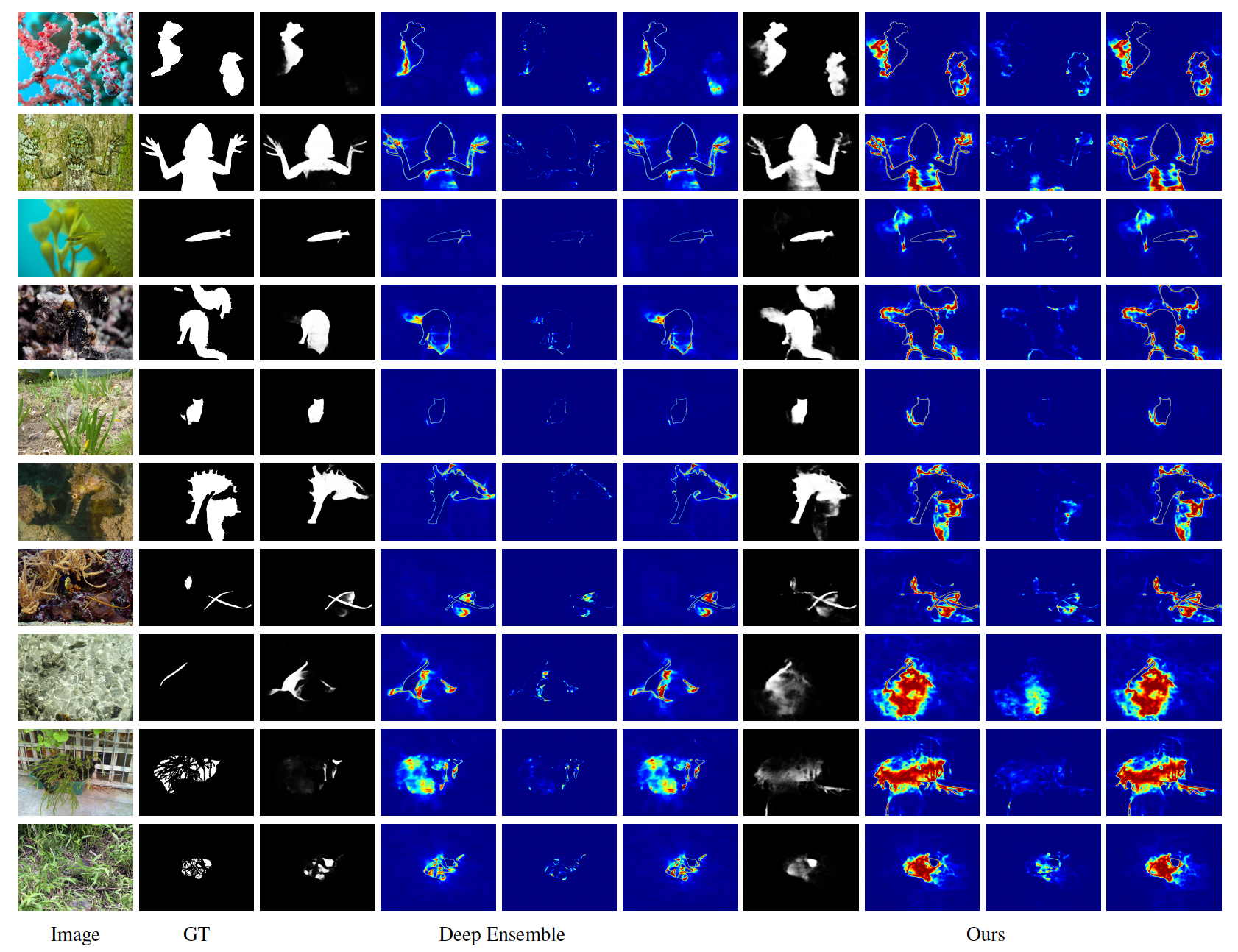}\\
\end{tabular}
  \end{center}
  \caption{Predictions and the generated uncertainty maps for camouflaged object detection. For the compared method and our method of each section, from left to right is: prediction, aleatoric uncertainty, epistemic uncertainty and predictive uncertainty.}
\label{fig:uncertainty_visualization_cod}
\end{figure*}

As a latent variable model, we also compare with alternative latent variable models, namely GAN \cite{gan_raw} and VAE \cite{VAE_Kingma,cvae}. Specifically, we define the
dimension of the latent variable $z$ same as ours within the
\enquote{GAN} and \enquote{VAE}.
For the GAN based model, an
additional discriminator is designed to distinguish the prediction and ground truth. For the VAE based framework, we use an
extra inference model $p(\vz \vert \vx,y)$ to approximate the true posterior distribution, and define $p(\vz \vert \vx)$ as the
prior of the latent variable following \cite{jing2020uc}. Similarly, we can obtain the three types of uncertainty with 10 forward passes at test time.

\noindent\textbf{GAN Implementation Details}
Within the GAN based framework, we design an extra fully convolutional discriminator $g_\beta$ following \cite{hung2018adversarial}, where $\beta$ is the parameter set of the discriminator.
Two different modules (the generator $f_\theta$ and the discriminator $g_\beta$ in our case) play the minimax game in GAN based framework:
\begin{equation}
\label{gan_loss}
\begin{aligned}
    \underset{f_\theta}{\min} \, \underset{g_\beta}{\max} \, V(g_\beta,f_\theta) &= E_{(\vx,y)\sim p_{data}(\vx,y)}[\log g_\beta(y|\vx)]\\ &+ E_{\vz\sim p(\vz)}[\log(1-g_\beta(f_\theta(\vx,\vz)))],
\end{aligned}
\end{equation}
where
$p_{data}(\vx,y)$ is the joint distribution of training data, $p(\vz)$ is the prior distribution of the latent variable $\vz$, which is usually defined as $p(\vz)=\mathcal{N}(0,\mathbf{I})$. In practice, we define loss function for the generator as the sum of a reconstruction loss $\mathcal{L}_{\text{rec}}$, and an adversarial loss $\mathcal{L}_{\text{adv}} = \mathcal{L}_{ce}(g_\beta(f_\theta(\vx,\vz)),1)$, which is $\mathcal{L}_{gen} = \mathcal{L}_{\text{rec}} + \lambda\mathcal{L}_{\text{adv}}$,
where the hyper-parameter $\lambda$ is tuned, and empirically we set $\lambda=0.1$ for stable training. $\mathcal{L}_{ce}$ is the binary cross-entropy loss.
The discriminator $g_\beta$ is trained via loss function as: $\mathcal{L}_{dis}=\mathcal{L}_{ce}(g_\beta(f_\theta(\vx,\vz)),0)+\mathcal{L}_{ce}(g_\beta(y),1)$.

We have almost the same structure for the prior net and the posterior net, where the only difference is that the input channel of the first convolutional layer within the prior net is 3 (channel of the RGB image $\vx$) and that within the posterior net is 4 (channel of the concatenation of $\vx$ and $y$).
We construct our prior and posterior net with five convolution layers of the same kernel size ($4\times 4$) and two fully connected layers. The channel size of the convolutional layers are $C'=C,2*C,4*C,8*C,8*C$ ($C=32$, which is the same as the channel reduction layers for the backbone features),
and we have a
batch normalization layer and a ReLU activation layer after each convolutional layer. The two fully connected layers are used to map the feature map to vectors representing the mean $\mu$ and standard deviation $\sigma$ of the latent variable $\vz$. The latent variable $\vz$ is then obtained via the reparameterization trick: $\vz=\mu+\sigma\odot\epsilon$, where $\epsilon\sim\mathcal{N}(0,\mathbf{I})$.

\noindent\textbf{VAE Implementation Details}
Given conditional variable $\vx$, we perform conditional variational auto-encoder (CVAE) \cite{cvae}. CVAE is a conditional directed graph model, which includes three variables, the input $x$ or conditional variable that modulates the prior on Gaussian latent variable $z$ and generates the output prediction $y$. Two main modules are included in a conventional CVAE based framework: a generator model $f_\theta(\vx,\vz)$
and an inference model $q_\theta(\vz|\vx,y)$, which infers the latent variable $\vz$ with image $\vx$ and annotation $y$ as input.
Learning a CVAEs framework involves approximation of the true posterior distribution of $\vz$ with an inference model $q_\theta(\vz|\vx,y)$, with the loss function as:
\begin{equation}
    \label{CVAE loss}
    \begin{aligned}
     \mathcal{L}_{cvae} = \underbrace{\mathbb{E}_{\vz\sim q_\theta(\vz|\vx,y)}[-{\rm log} p_\theta(y|\vx,\vz)]}_{\mathcal{L}_{\text{rec}}}\\
     + {D_{KL}}(q_\theta(\vz|\vx,y)\parallel p_\theta(\vz|\vx)),\\
    \end{aligned}
\end{equation}
where the first term is the reconstruction loss and the second is the Kullback-Leibler Divergence of prior distribution $p_\theta(\vz|\vx)$ and posterior distribution $q_\theta(\vz|\vx,y)$.

\begin{table*}[t!]
  \centering
  \scriptsize
  \renewcommand{\arraystretch}{1.2}
  \renewcommand{\tabcolsep}{0.3mm}
  \caption{Comprehensively analysis of the proposed single-pass ensemble based latent variable model for uncertainty estimation.}
  \begin{tabular}{lcccccc|cccc|cccc|cccc|cccc}
  \toprule
   Base&ABP&$p(z|x)$&DE&$U^a$&$U^{a+}$&$U^{p+}$ &\multicolumn{4}{c|}{CAMO~\cite{le2019anabranch}}&\multicolumn{4}{c|}{CHAMELEMON~\cite{Chameleon2018}}&\multicolumn{4}{c|}{COD10K~\cite{fan2020camouflaged}}&\multicolumn{4}{c}{NC4K~\cite{yunqiu2021ranking}} \\
    &&&&&&&$F_{\beta}\uparrow$&$\mathcal{M}\downarrow$&$\text{ECE}_d\downarrow$&$\text{PAvPU}\uparrow$
    &$F_{\beta}\uparrow$&$\mathcal{M}\downarrow$&$\text{ECE}_d\downarrow$&$\text{PAvPU}\uparrow$&$F_{\beta}\uparrow$&$\mathcal{M}\downarrow$&$\text{ECE}_d\downarrow$&$\text{PAvPU}\uparrow$ &$F_{\beta}\uparrow$&$\mathcal{M}\downarrow$&$\text{ECE}_d\downarrow$&$\text{PAvPU}\uparrow$\\
  \hline
  $\checkmark$&&&&&&& .736 & .083 & .066 & .838 & .832 & .033 & .020 & .875 & .710 & .038 & .029 & .876 & .785 & .053  & .041 & .851  \\
  $\checkmark$&$\checkmark$&&&&&& .731 & .086 & .057 & .870 & .814 & .036 & .018 & .908 & .694 & .041 & .024 & .904 & .774 & .054  & .034 & .884 \\
 $\checkmark$&$\checkmark$&$\checkmark$&&&&& .747 & .084 & .054 & .878 & .829 & .034 & .017 & .912 & .712 & .038 & .023 & .909 & .787 & .052 & .032 & .894 \\
  $\checkmark$&$\checkmark$&$\checkmark$&$\checkmark$&&&& .731 & .086 & .053 & .883 & .827 & .036 & .017 & .919 & .708 & .038 & .022 & .916 & .781 & .055 & .031 & .896 \\
  $\checkmark$&$\checkmark$&$\checkmark$&$\checkmark$&$\checkmark$&&& .737 & .087 & .051 & .886 & .821 & .037 & .015 & .921 & .709 & .039 & .020 & .920 & .780 & .054 & .030 & .900 \\
  $\checkmark$&$\checkmark$&$\checkmark$&$\checkmark$&$\checkmark$&$\checkmark$&& .736 & .084 & .047 & .894 & .828 & .034 & .013 & .929 & .713 & .037 & .016 & .927 & .778 & .054 & .026 & .911  \\
    $\checkmark$&$\checkmark$&$\checkmark$&$\checkmark$&$\checkmark$&$\checkmark$&$\checkmark$& .739 & .082 & .044 & .902 & .830 & .034 & .011 & .937 & .713 & .037 & .013 & .929 & .780 & .054 & .024 & .913 \\
   \bottomrule
  \end{tabular}
  \label{tab:ablation_total}
\end{table*}

Note that, the predictive uncertainty and aleatoric uncertainty of our method are obtained directly from the network, and the epistemic uncertainty is defined as their residual. The performance of the
MC-dropout, Deep ensemble, GAN and VAE based frameworks is shown in Table~\ref{tab:alternative_uncertainty_methods}, where the performance is
based on the mean prediction, and our performance is based on the single pass prediction from $f_\theta(\vx)$.
We observe comparable deterministic model performance of all the related solutions. However,
the significant lower $\text{ECE}_d$ and higher $\text{PAvPU}$ of our method validates the
superiority of our method in achieving a well-calibrated model.
We also visualize the produced uncertainty maps of deep ensemble model (\enquote{Deep Ensemble})
and ours
(\enquote{Ours}) in Fig.~\ref{fig:uncertainty_visualization_cod}, which explains the better uncertainty exploration of our solution in discovering the highly uncertain regions.

\subsubsection{Ablation Study}
With the baseline model (\enquote{Base}) in \Cref{tab:ablation_total},
we gradually add the proposed strategy to \enquote{Base} to extensively analyse our solutions, and show results in \Cref{tab:ablation_total}. Note that, for all the related models without the deterministic prediction branch ($f_\theta(\vx)$ in \Cref{tab:ablation_total}), the performance and the uncertainty measure are based on the mean model predictions with the same hyper-parameters as our main experiment. Otherwise,
the performance is based on $f_\theta(\vx)$.

Adding \enquote{ABP} to \enquote{Base} leading to an alternating back-propagation based latent variable model, where $p(\vz)=\mathcal{N}(0,\mathbf{I})$.
We introduce an
image-conditioned prior
$p(\vz \vert \vx)$,
and we call it the conditional alternating back-propagation (cABP). Based on cABP, we extend the decoder to a last-layer ensemble structure (\enquote{DE} in \Cref{tab:ablation_total}) with $M=5$ different output layers to generate $M$ different predictions.
Both the generative model based uncertainty estimation frameworks and the ensemble based frameworks usually obtain reliable uncertainty with decreased performance as a sacrifice \cite{kendall2018multi,Lu_2021_ICCV} (see  \Cref{tab:alternative_uncertainty_methods}). To maintain the deterministic prediction, we further introduce deterministic prediction branch $f_\theta(\vx)$ to \enquote{DE}, and obtain the
single-pass uncertainty estimation with uncertainty estimation modules ($U^a$, $U^{a+}$ and $U^{p+}$). With $U^a$, we ignore the \enquote{oracle} assumption, and define aleatoric uncertainty as the mean entropy of multiple predictions. With $U^{a+}$, the \enquote{oracle} assumption is taken into consideration, and we define uncertainty of the optimal prediction $f^\ast_\theta(\vx,\vz)$ as aleatoric uncertainty. For both $U^{a}$ and $U^{a+}$, the predictive uncertainty is only related to the uncertainty of the mean prediction.
With $U^{p+}$, we introduce $\text{err}$ as the second part of predictive uncertainty (see \cref{predictive_uncertainty_consistency}). The relatively comparable deterministic performance of $U^a$, $U^{a+}$, $U^{p+}$ and \enquote{Base} is consistent with one of our motivation that is to maintain the deterministic performance. The decreased $\text{ECE}_d$ and increased $\text{PAvPU}$ validate our solution in achieving a well-calibrated model, leading to comparable model accuracy and model confidence.

\subsection{Discussion}
According to the definition of aleatoric uncertainty and epistemic uncertainty, the former is inherent noise related, which cannot be explained away with more data, and the latter is knowledge gap driven, which can usually be reduced with more training samples. Based on this, given an already collected dataset, the inherent noise level is then fixed. The uncertainty difference with different ratios of labeled training dataset should mainly come from the epistemic uncertainty, making it promising to explore epistemic uncertainty and it's contribution in selecting diverse and representative samples for effective semi-supervised or active learning.

\section{Conclusion}
We thoroughly analysed
uncertainty estimation techniques with three important observations.
1) the trivial solution within dual-head based aleatoric uncertainty estimation framework \cite{nicta_uncertainty,kendall2017uncertainties}; 2) the hidden \enquote{oracle assumption} behind the aleatoric uncertainty definition \cite{deup_arxiv}; and
3) the latent variable within a conditional latent variable model as another origin of epistemic uncertainty. With the
above observations, we introduce a single-pass ensemble-based latent variable model to generate both accurate model prediction and reliable uncertainty. We apply our solution to a dense prediction task, namely camouflaged object detection.
Our extensive discussion and experimental results show that
the existing aleatoric uncertainty estimation techniques
have limitations in representing the \enquote{inherent noise} attribute.
With
exploration of the hidden \enquote{oracle} assumption, we are approximating the true aleatoric uncertainty. 
Further, we explain that latent variable within a conditional latent variable model can act as
another origin of epistemic uncertainty, leading to more comprehensive predictive uncertainty estimation.
{\small
\bibliographystyle{ieee_fullname}
\bibliography{egbib}

\begin{thebibliography}{10}\itemsep=-1pt

\bibitem{aitchison2021a}
Laurence Aitchison.
\newblock A statistical theory of cold posteriors in deep neural networks.
\newblock In {\em International Conference on Learning Representations (ICLR)},
  2021.

\bibitem{weight_uncertainty_nn}
Charles Blundell, Julien Cornebise, Koray Kavukcuoglu, and Daan Wierstra.
\newblock Weight uncertainty in neural network.
\newblock In {\em International Conference on Machine Learning (ICML)}, pages
  1613--1622, 2015.

\bibitem{explaining_bnn}
Kirill Bykov, Marina~M.{-}C. H{\"{o}}hne, Adelaida Creosteanu, Klaus{-}Robert
  M{\"{u}}ller, Frederick Klauschen, Shinichi Nakajima, and Marius Kloft.
\newblock Explaining bayesian neural networks.
\newblock {\em CoRR}, abs/2108.10346, 2021.

\bibitem{Carvalho_2020_CVPR}
Eduardo D.~C. Carvalho, Ronald Clark, Andrea Nicastro, and Paul H.~J. Kelly.
\newblock Scalable uncertainty for computer vision with functional variational
  inference.
\newblock In {\em IEEE Conference on Computer Vision and Pattern Recognition
  (CVPR)}, 2020.

\bibitem{Chatterjee_2021_ICCV}
Moitreya Chatterjee, Narendra Ahuja, and Anoop Cherian.
\newblock A hierarchical variational neural uncertainty model for stochastic
  video prediction.
\newblock In {\em IEEE International Conference on Computer Vision (ICCV)},
  pages 9751--9761, 2021.

\bibitem{bridging_the_gap_mcmc}
Changyou Chen, David Carlson, Zhe Gan, Chunyuan Li, and Lawrence Carin.
\newblock Bridging the gap between stochastic gradient mcmc and stochastic
  optimization.
\newblock In {\em Proceedings of the 19th International Conference on
  Artificial Intelligence and Statistics}, pages 1051--1060, 2016.

\bibitem{weakly_supervised_localization_uncertainty}
Xi Chen, Andy~J Ma, Nanxi Guo, and Jiajia Chen.
\newblock Improving weakly supervised object localization by uncertainty
  estimation of pseudo supervision.
\newblock In {\em IEEE International Conference on Multimedia and Expo (ICME)},
  pages 1--6, 2021.

\bibitem{Chitta2018DeepPE}
Kashyap Chitta, Jos{\'e}~Manuel {\'A}lvarez, and Adam Lesnikowski.
\newblock Deep probabilistic ensembles: Approximate variational inference
  through kl regularization.
\newblock {\em ArXiv}, abs/1811.02640, 2018.

\bibitem{Collier2020ASP}
Mark Collier, Basil Mustafa, Efi Kokiopoulou, Rodolphe Jenatton, and Jesse
  Berent.
\newblock A simple probabilistic method for deep classification under
  input-dependent label noise.
\newblock {\em arXiv: Learning}, 2020.

\bibitem{Collier_2021_CVPR}
Mark Collier, Basil Mustafa, Efi Kokiopoulou, Rodolphe Jenatton, and Jesse
  Berent.
\newblock Correlated input-dependent label noise in large-scale image
  classification.
\newblock In {\em IEEE Conference on Computer Vision and Pattern Recognition
  (CVPR)}, pages 1551--1560, 2021.

\bibitem{Depeweg2018DecompositionOU}
Stefan Depeweg, Jos{\'e}~Miguel Hern{\'a}ndez-Lobato, Finale Doshi-Velez, and
  S. Udluft.
\newblock Decomposition of uncertainty in bayesian deep learning for efficient
  and risk-sensitive learning.
\newblock In {\em International Conference on Machine Learning (ICML)}, 2018.

\bibitem{Durasov_2021_CVPR}
Nikita Durasov, Timur Bagautdinov, Pierre Baque, and Pascal Fua.
\newblock Masksembles for uncertainty estimation.
\newblock In {\em IEEE Conference on Computer Vision and Pattern Recognition
  (CVPR)}, pages 13539--13548, 2021.

\bibitem{Ekmekci_2021_ICCV}
Canberk Ekmekci and Mujdat Cetin.
\newblock What does your computational imaging algorithm not know?: A
  plug-and-play model quantifying model uncertainty.
\newblock In {\em IEEE International Conference on Computer Vision (ICCV)
  Workshop}, pages 4018--4027, 2021.

\bibitem{fan2020camouflaged}
Deng-Ping Fan, Ge-Peng Ji, Guolei Sun, Ming-Ming Cheng, Jianbing Shen, and Ling
  Shao.
\newblock Camouflaged object detection.
\newblock In {\em IEEE Conference on Computer Vision and Pattern Recognition
  (CVPR)}, pages 2777--2787, 2020.

\bibitem{labelnotperfect}
Di Feng, Zining Wang, Yiyang Zhou, Lars Rosenbaum, Fabian Timm, Klaus
  Dietmayer, Masayoshi Tomizuka, and Wei Zhan.
\newblock Labels are not perfect: Inferring spatial uncertainty in object
  detection.
\newblock {\em IEEE Transactions on Intelligent Transportation Systems}, pages
  1--14, 2021.

\bibitem{dropout_bayesian}
Yarin Gal and Zoubin Ghahramani.
\newblock Dropout as a bayesian approximation: Representing model uncertainty
  in deep learning.
\newblock In {\em International Conference on Machine Learning (ICML)}, pages
  1050--1059, 2016.

\bibitem{lightweight_probabilistic_network}
Jochen Gast and Stefan Roth.
\newblock Lightweight probabilistic deep networks.
\newblock In {\em IEEE Conference on Computer Vision and Pattern Recognition
  (CVPR)}, 2018.

\bibitem{gan_raw}
Ian Goodfellow, Jean Pouget-Abadie, Mehdi Mirza, Bing Xu, David Warde-Farley,
  Sherjil Ozair, Aaron Courville, and Yoshua Bengio.
\newblock Generative adversarial nets.
\newblock In {\em Conference on Neural Information Processing Systems (NIPS)},
  pages 2672--2680, 2014.

\bibitem{NIPS2014_5423_gan}
Ian Goodfellow, Jean Pouget-Abadie, Mehdi Mirza, Bing Xu, David Warde-Farley,
  Sherjil Ozair, Aaron Courville, and Yoshua Bengio.
\newblock Generative adversarial nets.
\newblock In {\em Conference on Neural Information Processing Systems (NIPS)},
  pages 2672--2680, 2014.

\bibitem{Groenendijk_2021_WACV}
Rick Groenendijk, Sezer Karaoglu, Theo Gevers, and Thomas Mensink.
\newblock Multi-loss weighting with coefficient of variations.
\newblock In {\em IEEE Winter Conference on Applications of Computer Vision
  (WACV)}, pages 1469--1478, 2021.

\bibitem{on_calibration}
Chuan Guo, Geoff Pleiss, Yu Sun, and Kilian~Q. Weinberger.
\newblock On calibration of modern neural networks.
\newblock In {\em International Conference on Machine Learning (ICML)}, pages
  1321--1330, 2017.

\bibitem{ABP_aaai}
Tian Han, Yang Lu, Song Zhu, and Yingnian Wu.
\newblock Alternating back-propagation for generator network.
\newblock In {\em AAAI Conference on Artificial Intelligence (AAAI)}, pages
  1976--1984, 2017.

\bibitem{mcmc_sampling}
W.~K. Hastings.
\newblock {Monte Carlo sampling methods using Markov chains and their
  applications}.
\newblock {\em Biometrika}, 57(1):97--109, 1970.

\bibitem{sampling_free_variation_inference}
Manuel Hau{\ss}mann, Fred~A. Hamprecht, and Melih Kandemir.
\newblock Sampling-free variational inference of bayesian neural networks by
  variance backpropagation.
\newblock In {\em Proceedings of The 35th Uncertainty in Artificial
  Intelligence Conference}, pages 563--573, 2020.

\bibitem{Lagging_Inference_Networks}
Junxian He, Daniel Spokoyny, Graham Neubig, and Taylor Berg-Kirkpatrick.
\newblock Lagging inference networks and posterior collapse in variational
  autoencoders.
\newblock In {\em International Conference on Learning Representations (ICLR)},
  2019.

\bibitem{he2016deep}
Kaiming He, Xiangyu Zhang, Shaoqing Ren, and Jian Sun.
\newblock Deep residual learning for image recognition.
\newblock In {\em IEEE Conference on Computer Vision and Pattern Recognition
  (CVPR)}, pages 770--778, 2016.

\bibitem{ResHe2015}
Kaiming He, Xiangyu Zhang, Shaoqing Ren, and Jian Sun.
\newblock Deep residual learning for image recognition.
\newblock In {\em IEEE Conference on Computer Vision and Pattern Recognition
  (CVPR)}, pages 770--778, 2016.

\bibitem{keep_nettwork_simple}
Geoffrey~E. Hinton and Drew van Camp.
\newblock Keeping the neural networks simple by minimizing the description
  length of the weights.
\newblock In {\em Proceedings of the Sixth Annual Conference on Computational
  Learning Theory}, page 5–13, 1993.

\bibitem{hung2018adversarial}
Wei-Chih Hung, Yi-Hsuan Tsai, Yan-Ting Liou, Yen-Yu Lin, and Ming-Hsuan Yang.
\newblock Adversarial learning for semi-supervised semantic segmentation.
\newblock {\em arXiv preprint arXiv:1802.07934}, 2018.

\bibitem{aleatoric_epistemic_concept}
Eyke Hüllermeier and Willem Waegeman.
\newblock Aleatoric and epistemic uncertainty in machine learning: an
  introduction to concepts and methods.
\newblock {\em Machine Learning}, 110, 03 2021.

\bibitem{what_are_bayesian_looklike}
Pavel Izmailov, Sharad Vikram, Matthew~D Hoffman, and Andrew Gordon~Gordon
  Wilson.
\newblock What are bayesian neural network posteriors really like?
\newblock In {\em International Conference on Machine Learning (ICML)}, pages
  4629--4640, 2021.

\bibitem{deup_arxiv}
Moksh Jain, Salem Lahlou, Hadi Nekoei, Victor Butoi, Paul Bertin, Jarrid
  Rector{-}Brooks, Maksym Korablyov, and Yoshua Bengio.
\newblock {DEUP:} direct epistemic uncertainty prediction.
\newblock {\em CoRR}, abs/2102.08501, 2021.

\bibitem{Jordan99anintroduction}
Michael~I. Jordan, Zoubin Ghahramani, and et al.
\newblock An introduction to variational methods for graphical models.
\newblock In {\em Machine Learning}, pages 183--233. MIT Press, 1999.

\bibitem{kendall2017uncertainties}
Alex Kendall and Yarin Gal.
\newblock What uncertainties do we need in bayesian deep learning for computer
  vision?
\newblock {\em arXiv preprint arXiv:1703.04977}, 2017.

\bibitem{kendall2018multi}
Alex Kendall, Yarin Gal, and Roberto Cipolla.
\newblock Multi-task learning using uncertainty to weigh losses for scene
  geometry and semantics.
\newblock In {\em IEEE Conference on Computer Vision and Pattern Recognition
  (CVPR)}, pages 7482--7491, 2018.

\bibitem{VAE_Kingma}
Diederik Kingma and Max Welling.
\newblock Auto-encoding variational bayes.
\newblock In {\em International Conference on Learning Representations (ICLR)},
  2014.

\bibitem{KIUREGHIAN2009105}
Armen~Der Kiureghian and Ove Ditlevsen.
\newblock Aleatory or epistemic? does it matter?
\newblock {\em Structural Safety}, 31(2):105--112, 2009.

\bibitem{equipping_dnn_uncertainty}
Lingkai Kong, Jimeng Sun, and Chao Zhang.
\newblock {SDE}-net: Equipping deep neural networks with uncertainty estimates.
\newblock In {\em International Conference on Machine Learning (ICML)}, pages
  5405--5415, 2020.

\bibitem{quatification_uncertainty}
Yongchan Kwon, Joong-Ho Won, Beom Kim, and Myunghee Paik.
\newblock Uncertainty quantification using bayesian neural networks in
  classification: Application to ischemic stroke lesion segmentation.
\newblock In {\em Medical Imaging with Deep Learning}, 04 2018.

\bibitem{kuppers2021bayesian}
Fabian Küppers, Jan Kronenberger, Jonas Schneider, and Anselm Haselhoff.
\newblock Bayesian confidence calibration for epistemic uncertainty modelling,
  2021.

\bibitem{simple_scalable_uncertainty}
Balaji Lakshminarayanan, Alexander Pritzel, and Charles Blundell.
\newblock Simple and scalable predictive uncertainty estimation using deep
  ensembles.
\newblock In {\em Conference on Neural Information Processing Systems (NIPS)},
  pages 6402--6413, 2017.

\bibitem{Well_Calibrated_Regression}
Max-Heinrich Laves, Sontje Ihler, Jacob~F. Fast, L\"uder~A. Kahrs, and Tobias
  Ortmaier.
\newblock Well-calibrated regression uncertainty in medical imaging with deep
  learning.
\newblock In {\em Proceedings of the Third Conference on Medical Imaging with
  Deep Learning}, pages 393--412, 2020.

\bibitem{le2019anabranch}
Trung-Nghia Le, Tam~V Nguyen, Zhongliang Nie, Minh-Triet Tran, and Akihiro
  Sugimoto.
\newblock Anabranch network for camouflaged object segmentation.
\newblock {\em Computer Vision and Image Understanding}, 184:45--56, 2019.

\bibitem{Li_2021_CVPR_ordinal_embedding}
Wanhua Li, Xiaoke Huang, Jiwen Lu, Jianjiang Feng, and Jie Zhou.
\newblock Learning probabilistic ordinal embeddings for uncertainty-aware
  regression.
\newblock In {\em IEEE Conference on Computer Vision and Pattern Recognition
  (CVPR)}, pages 13896--13905, 2021.

\bibitem{Liu2020SimpleAP}
Jeremiah~Zhe Liu, Zi Lin, Shreyas Padhy, Dustin Tran, Tania Bedrax-Weiss, and
  Balaji Lakshminarayanan.
\newblock Simple and principled uncertainty estimation with deterministic deep
  learning via distance awareness.
\newblock In {\em Conference on Neural Information Processing Systems
  (NeurIPS)}, 2020.

\bibitem{LIU2021108140}
Lu Liu and Robby~T. Tan.
\newblock Certainty driven consistency loss on multi-teacher networks for
  semi-supervised learning.
\newblock {\em Pattern Recognition}, 120:108140, 2021.

\bibitem{Long_2015_CVPR}
Jonathan Long, Evan Shelhamer, and Trevor Darrell.
\newblock Fully convolutional networks for semantic segmentation.
\newblock In {\em IEEE Conference on Computer Vision and Pattern Recognition
  (CVPR)}, 2015.

\bibitem{Lu_2021_ICCV}
Yan Lu, Xinzhu Ma, Lei Yang, Tianzhu Zhang, Yating Liu, Qi Chu, Junjie Yan, and
  Wanli Ouyang.
\newblock Geometry uncertainty projection network for monocular 3d object
  detection.
\newblock In {\em IEEE International Conference on Computer Vision (ICCV)},
  pages 3111--3121, 2021.

\bibitem{yunqiu2021ranking}
Yunqiu Lyu, Jing Zhang, Yuchao Dai, Aixuan Li, Bowen Liu, Nick Barnes, and
  Deng-Ping Fan.
\newblock Simultaneously localize, segment and rank the camouflaged objects.
\newblock In {\em IEEE Conference on Computer Vision and Pattern Recognition
  (CVPR)}, 2021.

\bibitem{Mirikharaji_2021_CVPR}
Zahra Mirikharaji, Kumar Abhishek, Saeed Izadi, and Ghassan Hamarneh.
\newblock D-lema: Deep learning ensembles from multiple annotations -
  application to skin lesion segmentation.
\newblock In {\em IEEE Conference on Computer Vision and Pattern Recognition
  (CVPR) Workshop}, pages 1837--1846, 2021.

\bibitem{evaluation_uncertainty}
Jishnu Mukhoti and Yarin Gal.
\newblock Evaluating bayesian deep learning methods for semantic segmentation.
\newblock {\em CoRR}, abs/1811.12709, 2018.

\bibitem{mukhoti2021deterministic}
Jishnu Mukhoti, Andreas Kirsch, Joost van Amersfoort, Philip~HS Torr, and Yarin
  Gal.
\newblock Deterministic neural networks with appropriate inductive biases
  capture epistemic and aleatoric uncertainty.
\newblock {\em arXiv preprint arXiv:2102.11582}, 2021.

\bibitem{mcmc_langevin}
Radford Neal.
\newblock Mcmc using hamiltonian dynamics.
\newblock {\em Handbook of Markov Chain Monte Carlo}, 06 2012.

\bibitem{bayesian_learning_for_nn}
Radford~M. Neal.
\newblock {\em Bayesian Learning for Neural Networks}.
\newblock Springer-Verlag, 1996.

\bibitem{Nguyen_2021_ICCV}
Khoi Nguyen and Sinisa Todorovic.
\newblock A weakly supervised amodal segmenter with boundary uncertainty
  estimation.
\newblock In {\em IEEE International Conference on Computer Vision (ICCV)},
  pages 7396--7405, October 2021.

\bibitem{nicta_uncertainty}
D.A. Nix and A.S. Weigend.
\newblock Estimating the mean and variance of the target probability
  distribution.
\newblock In {\em Proceedings of 1994 IEEE International Conference on Neural
  Networks}, 1994.

\bibitem{global_inducing_point_variational}
Sebastian~W Ober and Laurence Aitchison.
\newblock Global inducing point variational posteriors for bayesian neural
  networks and deep gaussian processes.
\newblock In {\em International Conference on Machine Learning (ICML)}, pages
  8248--8259, 2021.

\bibitem{epistemic_ian}
Ian Osband, Zheng Wen, Mohammad Asghari, Morteza Ibrahimi, Xiyuan Lu, and
  Benjamin~Van Roy.
\newblock Epistemic neural networks.
\newblock {\em CoRR}, abs/2107.08924, 2021.

\bibitem{Peng_2021_ICCV}
Fengchao Peng, Chao Wang, Jianzhuang Liu, and Zhen Yang.
\newblock Active learning for lane detection: A knowledge distillation
  approach.
\newblock In {\em IEEE International Conference on Computer Vision (ICCV)},
  pages 15152--15161, 2021.

\bibitem{Postels_2019_ICCV_Sampling_Free}
Janis Postels, Francesco Ferroni, Huseyin Coskun, Nassir Navab, and Federico
  Tombari.
\newblock Sampling-free epistemic uncertainty estimation using approximated
  variance propagation.
\newblock In {\em IEEE International Conference on Computer Vision (ICCV)},
  2019.

\bibitem{deterministic_epistemic}
Janis Postels, Mattia Seg{\`{u}}, Tao Sun, Luc~Van Gool, Fisher Yu, and
  Federico Tombari.
\newblock On the practicality of deterministic epistemic uncertainty.
\newblock {\em CoRR}, abs/2107.00649, 2021.

\bibitem{midas_tpami}
Rene Ranftl, Katrin Lasinger, David Hafner, Konrad Schindler, and Vladlen
  Koltun.
\newblock Towards robust monocular depth estimation: Mixing datasets for
  zero-shot cross-dataset transfer.
\newblock {\em IEEE Transactions on Pattern Analysis and Machine Intelligence
  (TPAMI)}, pages 1--1, 2020.

\bibitem{Rasmussen2004}
Carl~Edward Rasmussen.
\newblock {\em Gaussian Processes in Machine Learning}, pages 63--71.
\newblock Springer Berlin Heidelberg, 2004.

\bibitem{S_2021_CVPR_domain}
Prabhu~Teja S and Francois Fleuret.
\newblock Uncertainty reduction for model adaptation in semantic segmentation.
\newblock In {\em IEEE Conference on Computer Vision and Pattern Recognition
  (CVPR)}, pages 9613--9623, 2021.

\bibitem{generating_counterfactual}
Lisa Schut, Oscar Key, Rory Mc~Grath, Luca Costabello, Bogdan Sacaleanu, Medb
  Corcoran, and Yarin Gal.
\newblock Generating interpretable counterfactual explanations by implicit
  minimisation of epistemic and aleatoric uncertainties.
\newblock In {\em Proceedings of The 24th International Conference on
  Artificial Intelligence and Statistics}, pages 1756--1764, 2021.

\bibitem{Chameleon2018}
P Skurowski, H Abdulameer, J B{\l}aszczyk, T Depta, A Kornacki, and P
  Kozie{\l}.
\newblock Animal camouflage analysis: Chameleon database.
\newblock In {\em Unpublished Manuscript}, 2018.

\bibitem{cvae}
Kihyuk Sohn, Honglak Lee, and Xinchen Yan.
\newblock Learning structured output representation using deep conditional
  generative models.
\newblock In {\em Conference on Neural Information Processing Systems (NIPS)},
  pages 3483--3491, 2015.

\bibitem{Truong_2021_CVPR}
Prune Truong, Martin Danelljan, Luc Van~Gool, and Radu Timofte.
\newblock Learning accurate dense correspondences and when to trust them.
\newblock In {\em IEEE Conference on Computer Vision and Pattern Recognition
  (CVPR)}, pages 5714--5724, 2021.

\bibitem{Upadhyay_2021_ICCV}
Uddeshya Upadhyay, Viswanath~P. Sudarshan, and Suyash~P. Awate.
\newblock Uncertainty-aware gan with adaptive loss for robust mri image
  enhancement.
\newblock In {\em IEEE International Conference on Computer Vision (ICCV)
  Workshop}, pages 3255--3264, 2021.

\bibitem{vanamersfoort2021feature}
Joost van Amersfoort, Lewis Smith, Andrew Jesson, Oscar Key, and Yarin Gal.
\newblock On feature collapse and deep kernel learning for single forward pass
  uncertainty, 2021.

\bibitem{van2020uncertainty}
Joost van Amersfoort, Lewis Smith, Yee~Whye Teh, and Yarin Gal.
\newblock Uncertainty estimation using a single deep deterministic neural
  network.
\newblock In {\em International Conference on Machine Learning (ICML)}, 2020.

\bibitem{Graphical_Models_VI}
Martin~J. Wainwright and Michael~I. Jordan.
\newblock Graphical models, exponential families, and variational inference.
\newblock {\em Foundations and Trends® in Machine Learning}, 1(1–2):1--305,
  2008.

\bibitem{latent_derivative_bayesian}
Joe Watson, Jihao Andreas~Lin, Pascal Klink, Joni Pajarinen, and Jan Peters.
\newblock Latent derivative bayesian last layer networks.
\newblock In {\em Proceedings of The 24th International Conference on
  Artificial Intelligence and Statistics}, pages 1198--1206, 2021.

\bibitem{wei2020f3net}
Jun Wei, Shuhui Wang, and Qingming Huang.
\newblock F$^3$net: Fusion, feedback and focus for salient object detection.
\newblock In {\em AAAI Conference on Artificial Intelligence (AAAI)}, pages
  12321--12328, 2020.

\bibitem{batchensemble}
Yeming Wen, Dustin Tran, and Jimmy Ba.
\newblock Batchensemble: an alternative approach to efficient ensemble and
  lifelong learning.
\newblock In {\em International Conference on Learning Representations (ICLR)},
  2020.

\bibitem{multi_view_uncertainty}
Yingda Xia, Dong Yang, Zhiding Yu, Fengze Liu, Jinzheng Cai, Lequan Yu, Zhuotun
  Zhu, Daguang Xu, Alan~L. Yuille, and Holger Roth.
\newblock Uncertainty-aware multi-view co-training for semi-supervised medical
  image segmentation and domain adaptation.
\newblock In {\em AAAI Conference on Artificial Intelligence (AAAI)}, 2021.

\bibitem{random_relu}
Yufeng Xia, Jun Zhang, Zhiqiang Gong, Tingsong Jiang, and Wen Yao.
\newblock Randomized relu activation for uncertainty estimation of deep neural
  networks.
\newblock {\em CoRR}, abs/2107.07197, 2021.

\bibitem{Xu_2021_ICCV_mutltiview}
Hongbin Xu, Zhipeng Zhou, Yali Wang, Wenxiong Kang, Baigui Sun, Hao Li, and Yu
  Qiao.
\newblock Digging into uncertainty in self-supervised multi-view stereo.
\newblock In {\em IEEE International Conference on Computer Vision (ICCV)},
  pages 6078--6087, 2021.

\bibitem{denseaspp}
Maoke Yang, Kun Yu, Chi Zhang, Zhiwei Li, and Kuiyuan Yang.
\newblock Denseaspp for semantic segmentation in street scenes.
\newblock In {\em IEEE Conference on Computer Vision and Pattern Recognition
  (CVPR)}, pages 3684--3692, 2018.

\bibitem{Yang_2021_CVPR}
Wenfei Yang, Tianzhu Zhang, Xiaoyuan Yu, Tian Qi, Yongdong Zhang, and Feng Wu.
\newblock Uncertainty guided collaborative training for weakly supervised
  temporal action detection.
\newblock In {\em IEEE Conference on Computer Vision and Pattern Recognition
  (CVPR)}, pages 53--63, 2021.

\bibitem{advance_variational_inference}
C. Zhang, J. Butepage, H. Kjellstrom, and S. Mandt.
\newblock Advances in variational inference.
\newblock {\em IEEE Transactions on Pattern Analysis and Machine Intelligence
  (TPAMI)}, 41(08):2008--2026, 2019.

\bibitem{zhang2020uncertaintyaware}
Jing Zhang, Yuchao Dai, Xin Yu, Mehrtash Harandi, Nick Barnes, and Richard
  Hartley.
\newblock Uncertainty-aware deep calibrated salient object detection.
\newblock {\em arXiv preprint arXiv:2012.06020}, 2020.

\bibitem{jing2020uc}
Jing Zhang, Deng-Ping Fan, Yuchao Dai, Saeed Anwar, Fatemeh {Sadat Saleh}, Tong
  Zhang, and Nick Barnes.
\newblock Uc-net: Uncertainty inspired {RGB-D} saliency detection via
  conditional variational autoencoders.
\newblock In {\em IEEE Conference on Computer Vision and Pattern Recognition
  (CVPR)}, pages 8582--8591, 2020.

\bibitem{Zhuang_2021_CVPR}
Bingbing Zhuang and Manmohan Chandraker.
\newblock Fusing the old with the new: Learning relative camera pose with
  geometry-guided uncertainty.
\newblock In {\em IEEE Conference on Computer Vision and Pattern Recognition
  (CVPR)}, pages 32--42, 2021.

\end{thebibliography}
}

\end{document}